\pdfoutput=1
\documentclass[letterpaper, 12 pt,oneside,fleqn]{article}  

\usepackage{graphicx} 
\usepackage{cite} 
\usepackage[caption=false,font=scriptsize,justification=raggedright,singlelinecheck=true]{subfig} 
\usepackage{verbatim}  
\usepackage{url} 
\usepackage[multidot]{grffile} 
\usepackage{amsmath} 
\usepackage{amssymb}  
\usepackage{glossaries}
\usepackage{sectsty}
\usepackage{times}
\usepackage{fancyhdr}
\usepackage[normalem]{ulem} 

\usepackage[usenames,dvipsnames]{xcolor}

\usepackage{multirow}
\usepackage{ctable}

\makeglossaries 

\usepackage{bm} 

\DeclareMathSymbol{\Gamma}{\mathalpha}{letters}{"00}
\DeclareMathSymbol{\Delta}{\mathalpha}{letters}{"01}
\DeclareMathSymbol{\Theta}{\mathalpha}{letters}{"02}
\DeclareMathSymbol{\Lambda}{\mathalpha}{letters}{"03}
\DeclareMathSymbol{\Xi}{\mathalpha}{letters}{"04}
\DeclareMathSymbol{\Pi}{\mathalpha}{letters}{"05}
\DeclareMathSymbol{\Sigma}{\mathalpha}{letters}{"06}
\DeclareMathSymbol{\Upsilon}{\mathalpha}{letters}{"07}
\DeclareMathSymbol{\Phi}{\mathalpha}{letters}{"08}
\DeclareMathSymbol{\Psi}{\mathalpha}{letters}{"09}
\DeclareMathSymbol{\Omega}{\mathalpha}{letters}{"0A}

\newcommand\sub[1]{_{\scriptscriptstyle\mathit{#1}}}

\DeclareSymbolFont{EUr}{U}{eur}{m}{n}
\DeclareSymbolFont{EUb}{U}{eur}{b}{n}
\DeclareMathSymbol{\upGamma}{\mathord}{EUr}{"00}
\DeclareMathSymbol{\upDelta}{\mathord}{EUr}{"01}
\DeclareMathSymbol{\upTheta}{\mathord}{EUr}{"02}
\DeclareMathSymbol{\upLambda}{\mathord}{EUr}{"03}
\DeclareMathSymbol{\upXi}{\mathord}{EUr}{"04}
\DeclareMathSymbol{\upPi}{\mathord}{EUr}{"05}
\DeclareMathSymbol{\upSigma}{\mathord}{EUr}{"06}
\DeclareMathSymbol{\upUpsilon}{\mathord}{EUr}{"07}
\DeclareMathSymbol{\upPhi}{\mathord}{EUr}{"08}
\DeclareMathSymbol{\upPsi}{\mathord}{EUr}{"09}
\DeclareMathSymbol{\upOmega}{\mathord}{EUr}{"0A}
\DeclareMathSymbol{\upalpha}{\mathord}{EUr}{"0B}
\DeclareMathSymbol{\upbeta}{\mathord}{EUr}{"0C}
\DeclareMathSymbol{\upgamma}{\mathord}{EUr}{"0D}
\DeclareMathSymbol{\updelta}{\mathord}{EUr}{"0E}
\DeclareMathSymbol{\upepsilon}{\mathord}{EUr}{"0F}
\DeclareMathSymbol{\upzeta}{\mathord}{EUr}{"10}
\DeclareMathSymbol{\upeta}{\mathord}{EUr}{"11}
\DeclareMathSymbol{\uptheta}{\mathord}{EUr}{"12}
\DeclareMathSymbol{\upiota}{\mathord}{EUr}{"13}
\DeclareMathSymbol{\upkappa}{\mathord}{EUr}{"14}
\DeclareMathSymbol{\uplambda}{\mathord}{EUr}{"15}
\DeclareMathSymbol{\upmu}{\mathord}{EUr}{"16}
\DeclareMathSymbol{\upnu}{\mathord}{EUr}{"17}
\DeclareMathSymbol{\upxi}{\mathord}{EUr}{"18}
\DeclareMathSymbol{\uppi}{\mathord}{EUr}{"19}
\DeclareMathSymbol{\uprho}{\mathord}{EUr}{"1A}
\DeclareMathSymbol{\upsigma}{\mathord}{EUr}{"1B}
\DeclareMathSymbol{\uptau}{\mathord}{EUr}{"1C}
\DeclareMathSymbol{\upupsilon}{\mathord}{EUr}{"1D}
\DeclareMathSymbol{\upphi}{\mathord}{EUr}{"1E}
\DeclareMathSymbol{\upchi}{\mathord}{EUr}{"1F}
\DeclareMathSymbol{\uppsi}{\mathord}{EUr}{"20}
\DeclareMathSymbol{\upomega}{\mathord}{EUr}{"21}
\DeclareMathSymbol{\upvarepsilon}{\mathord}{EUr}{"22}
\DeclareMathSymbol{\upvartheta}{\mathord}{EUr}{"23}
\DeclareMathSymbol{\upvaromega}{\mathord}{EUr}{"24}
\DeclareMathSymbol{\upvarphi}{\mathord}{EUr}{"27}
\DeclareMathSymbol{\UpGamma}{\mathord}{EUb}{"00}
\DeclareMathSymbol{\UpDelta}{\mathord}{EUb}{"01}
\DeclareMathSymbol{\UpTheta}{\mathord}{EUb}{"02}
\DeclareMathSymbol{\UpLambda}{\mathord}{EUb}{"03}
\DeclareMathSymbol{\UpXi}{\mathord}{EUb}{"04}
\DeclareMathSymbol{\UpPi}{\mathord}{EUb}{"05}
\DeclareMathSymbol{\UpSigma}{\mathord}{EUb}{"06}
\DeclareMathSymbol{\UpUpsilon}{\mathord}{EUb}{"07}
\DeclareMathSymbol{\UpPhi}{\mathord}{EUb}{"08}
\DeclareMathSymbol{\UpPsi}{\mathord}{EUb}{"09}
\DeclareMathSymbol{\UpOmega}{\mathord}{EUb}{"0A}
\DeclareMathSymbol{\Upalpha}{\mathord}{EUb}{"0B}
\DeclareMathSymbol{\Upbeta}{\mathord}{EUb}{"0C}
\DeclareMathSymbol{\Upgamma}{\mathord}{EUb}{"0D}
\DeclareMathSymbol{\Updelta}{\mathord}{EUb}{"0E}
\DeclareMathSymbol{\Upepsilon}{\mathord}{EUb}{"0F}
\DeclareMathSymbol{\Upzeta}{\mathord}{EUb}{"10}
\DeclareMathSymbol{\Upeta}{\mathord}{EUb}{"11}
\DeclareMathSymbol{\Uptheta}{\mathord}{EUb}{"12}
\DeclareMathSymbol{\Upiota}{\mathord}{EUb}{"13}
\DeclareMathSymbol{\Upkappa}{\mathord}{EUb}{"14}
\DeclareMathSymbol{\Uplambda}{\mathord}{EUb}{"15}
\DeclareMathSymbol{\Upmu}{\mathord}{EUb}{"16}
\DeclareMathSymbol{\Upnu}{\mathord}{EUb}{"17}
\DeclareMathSymbol{\Upxi}{\mathord}{EUb}{"18}
\DeclareMathSymbol{\Uppi}{\mathord}{EUb}{"19}
\DeclareMathSymbol{\Uprho}{\mathord}{EUb}{"1A}
\DeclareMathSymbol{\Upsigma}{\mathord}{EUb}{"1B}
\DeclareMathSymbol{\Uptau}{\mathord}{EUb}{"1C}
\DeclareMathSymbol{\Upupsilon}{\mathord}{EUb}{"1D}
\DeclareMathSymbol{\Upphi}{\mathord}{EUb}{"1E}
\DeclareMathSymbol{\Upchi}{\mathord}{EUb}{"1F}
\DeclareMathSymbol{\Uppsi}{\mathord}{EUb}{"20}
\DeclareMathSymbol{\Upomega}{\mathord}{EUb}{"21}
\DeclareMathSymbol{\Upvarepsilon}{\mathord}{EUb}{"22}
\DeclareMathSymbol{\Upvartheta}{\mathord}{EUb}{"23}
\DeclareMathSymbol{\Upvaromega}{\mathord}{EUb}{"24}
\DeclareMathSymbol{\Upvarphi}{\mathord}{EUb}{"27}

\newcount\ppnum
\newcommand\ppnumber[1]{%
        \ppnum=#1\relax
        \ifnum\ppnum<0
                $-$%
                \ppnum=-\ppnum
        \fi
        \let\pptemp\empty
        \loop\ifnum\ppnum>999
                \count255=\ppnum
                \divide\ppnum by1000
                \count255=\numexpr \count255 - 1000*\ppnum \relax
                \edef\pptemp{,\!\ifnum\count255<100 0\ifnum\count255<10 0\fi\fi
                             \the\count255 \pptemp}%
        \repeat
        \the\ppnum
        \pptemp
}
\newacronym{ACFR}{ACFR}{Australian centre for field robotics}
\newacronym{ACRV}{ACRV}{Australian centre for robotic vision}

\newacronym{AUV}{AUV}{autonomous underwater vehicle}
\newacronym{UAV}{UAV}{unmanned aerial vehicle}
\newacronym{USV}{USV}{unmanned surface vehicle}
\newacronym{UGV}{UGV}{unmanned ground vehicle}
\newacronym{GPS}{GPS}{global positioning system}
\newacronym{SLAM}{SLAM}{simultaneous localisation and mapping}

\newacronym{MDSP}{MDSP}{multi-dimensional signal processing}
\newacronym{ROS}{ROS}{region of support}
\newacronym{DOF}{DOF}{degree-of-freedom}
\newacronym{RMS}{RMS}{root mean square}
\newacronym{SNR}{SNR}{signal-to-noise ratio}
\newacronym{CNR}{CNR}{contrast-to-noise ratio}
\newacronym{PCA}{PCA}{principal component analysis}

\newacronym{FIR}{FIR}{finite impulse response}
\newacronym{IIR}{IIR}{infinite impulse response}
\newacronym{DFT}{DFT}{discrete Fourier transform}
\newacronym{FFT}{FFT}{fast Fourier transform}
\newacronym{PSNR}{PSNR}{peak signal-to-noise ratio}
\newacronym{FPGA}{FPGA}{field programmable gate array}
\newacronym{GPU}{GPU}{graphics processing unit}
\newacronym{ASIC}{ASIC}{application-specific integrated circuit}
\newacronym{BW}{BW}{bandwidth}

\newacronym{PSF}{PSF}{point spread function}
\newacronym{FOV}{FOV}{field of view}
\newacronym{BRDF}{BRDF}{bidirectional reflectance distribution function}
\newacronym{FWHM}{FWHM}{full width at half maximum}

\newacronym{RANSAC}{RANSAC}{random sampling and consensus}
\newacronym{IBVS}{IBVS}{image-based visual servoing}
\newacronym{PBVS}{PBVS}{position-based visual servoing}
\newacronym{VS}{VS}{visual servoing}
\newacronym{LF}{LF}{light field}
\newacronym{LF-IBVS}{LF-IBVS}{light field image-based visual servoing}
\newacronym{M-IBVS}{M-IBVS}{monocular image-based visual servoing}

\usepackage{xifthen}

\newcommand{\bi}{\begin{itemize}}
\newcommand{\ei}{\end{itemize}}

\newcommand{\bfig}{\begin{figure}}
\newcommand{\efig}{\end{figure}}

\newcommand{\be}{\begin{equation}}
\newcommand{\ee}{\end{equation}}

\newcommand{\ba}{\begin{eqnarray}}
\newcommand{\ea}{\end{eqnarray}}

\newcommand{\REF}[2][ZZZZ]{\ifthenelse{\equal{#1}{ZZZZ}}
  {\index{general}{#2}\ifthenelse{\boolean{draft}}{{\color{red}\it#2}}{#2}}
  {\index{general}{#1}\ifthenelse{\boolean{draft}}{{\color{red}\it#2}}{#2}}}

\newcommand{\DEF}[2][ZZZZ]{\ifthenelse{\equal{#1}{ZZZZ}}
  {\index{general}{#2}\ifthenelse{\boolean{draft}}{{\color{red}\it#2}}{#2}}
  {\index{general}{#1}\ifthenelse{\boolean{draft}}{{\color{red}\it#2}}{#2}}}

\newcommand{\DEFX}[2][ZZZZ]{\ifthenelse{\equal{#1}{ZZZZ}}
  {\index{general}{#2|textbf}\ifthenelse{\boolean{draft}}{{\color{red}\it#2}}{#2}}
  {\index{general}{#1|textbf}\ifthenelse{\boolean{draft}}{{\color{red}\it#2}}{#2}}}

\newcommand{\model}[1]{\index{code}{#1@\textit{#1}}\ifthenelse{\boolean{draft}}{{\color{green}\Verb+#1+}}{\Verb+#1+}}
\newcommand{\block}[1]{\ifthenelse{\boolean{draft}}{{\color{green}\Verb+#1+}}{\textsf{#1}}}

\newcommand{\func}[2][ZZZZ]{\ifthenelse{\equal{#1}{ZZZZ}}{\index{code}{#2}}{\index{code}{#1}}\ifthenelse{\boolean{draft}}{{\color{green}\Verb+#2+}}{\Verb+#2+}}

\newcommand{\methodb}[2]{\index{code}{#1@\textbf{#1}!.#2}\ifthenelse{\boolean{draft}}{{\color{magenta}\Verb+#1.#2+}}{\Verb+#1.#2+}}
\newcommand{\method}[2]{\index{code}{#1@\textbf{#1}!.#2}\ifthenelse{\boolean{draft}}{{\color{magenta}\Verb+#2+}}{\Verb+#2+}}
\newcommand{\class}[1]{\index{code}{#1@\textbf{#1}}\ifthenelse{\boolean{draft}}{{\color{cyan}\Verb+#1+}}{\Verb+#1+}}
\newcommand{\property}[1]{\index{property}{#1}\ifthenelse{\boolean{draft}}{{\color{cyan}\Verb+#1+}}{\Verb+#1+}}

\newcommand{\presup}[1]{\,{}^{\scriptscriptstyle #1}\!}

\newcommand{\pose}[1][ZZZZ]{\ifthenelse{\equal{#1}{ZZZZ}}{}{\presup{#1}}{\mathbf{\xi}}}

\usepackage{mathrsfs}

\newcommand{\poser}[1][ZZZZ]{\ifthenelse{\equal{#1}{ZZZZ}}{}{\presup{#1}}{\mathscr{R}}}
\newcommand{\poserx}[1][ZZZZ]{\ifthenelse{\equal{#1}{ZZZZ}}{}{\presup{#1}}{\mathscr{R}_x}}
\newcommand{\posery}[1][ZZZZ]{\ifthenelse{\equal{#1}{ZZZZ}}{}{\presup{#1}}{\mathscr{R}_y}}
\newcommand{\poserz}[1][ZZZZ]{\ifthenelse{\equal{#1}{ZZZZ}}{}{\presup{#1}}{\mathscr{R}_z}}
\newcommand{\poserw}[1][ZZZZ]{\ifthenelse{\equal{#1}{ZZZZ}}{}{\presup{#1}}{\mathscr{R}_\omega}}
\newcommand{\poset}[1][ZZZZ]{\ifthenelse{\equal{#1}{ZZZZ}}{}{\presup{#1}}{\mathscr{T}}}
\newcommand{\posetx}[1][ZZZZ]{\ifthenelse{\equal{#1}{ZZZZ}}{}{\presup{#1}}{\mathscr{T}_x}}
\newcommand{\posety}[1][ZZZZ]{\ifthenelse{\equal{#1}{ZZZZ}}{}{\presup{#1}}{\mathscr{T}_y}}
\newcommand{\posetz}[1][ZZZZ]{\ifthenelse{\equal{#1}{ZZZZ}}{}{\presup{#1}}{\mathscr{T}_z}}
\newcommand{\posett}[1][ZZZZ]{\ifthenelse{\equal{#1}{ZZZZ}}{}{\presup{#1}}{\mathscr{T}\!}}

\newcommand{\poseri}[1][ZZZZ]{\ifthenelse{\equal{#1}{ZZZZ}}{}{\presup{#1}}{\mathscr{R}_i}}
\newcommand{\poseti}[1][ZZZZ]{\ifthenelse{\equal{#1}{ZZZZ}}{}{\presup{#1}}{\mathscr{T}_i}}

\newcommand{\twist}[1][ZZZZ]{\ifthenelse{\equal{#1}{ZZZZ}}{}{\presup{#1}}{S}}

\newcommand{\estpose}[1][ZZZZ]{\ifthenelse{\equal{#1}{ZZZZ}}{}{\presup{#1}}{\mathbf{\hat{\xi}}}}
\newcommand{\hpose}[1][ZZZZ]{\ifthenelse{\equal{#1}{ZZZZ}}{}{\presup{#1}}{\hat{\mathbf{\xi}}}}
\newcommand{\posedot}[1][ZZZZ]{\ifthenelse{\equal{#1}{ZZZZ}}{}{\presup{#1}}{\mathbf{\nu}}}
\newcommand{\q}[1][ZZZZ]{\ifthenelse{\equal{#1}{ZZZZ}}{}{\presup{#1}}{\mathring{q}}}

\DeclareMathAlphabet{\mathitbf}{OML}{cmm}{b}{it}
\renewcommand{\vec}[2][ZZZZ]{\ifthenelse{\equal{#1}{ZZZZ}}{}{\presup{#1}}{\mathitbf{#2}}}

\newcommand{\hvec}[2][ZZZZ]{\ifthenelse{\equal{#1}{ZZZZ}}{}{\presup{#1}}{\hat{\vec{#2}}}}
\newcommand{\ovec}[2][ZZZZ]{\ifthenelse{\equal{#1}{ZZZZ}}{}{\presup{#1}}{\mathring{\vec{#2}}}}
\newcommand{\tvec}[2][ZZZZ]{\ifthenelse{\equal{#1}{ZZZZ}}{}{\presup{#1}}{\tilde{\vec{#2}}}}
\newcommand{\evec}[2][ZZZZ]{\ifthenelse{\equal{#1}{ZZZZ}}{}{\presup{#1}}{\hat{\vec{#2}}}}
\newcommand{\dvec}[2][ZZZZ]{\ifthenelse{\equal{#1}{ZZZZ}}{}{\presup{#1}}{\dot{\vec{#2}}}}
\newcommand{\ddvec}[2][ZZZZ]{\ifthenelse{\equal{#1}{ZZZZ}}{}{\presup{#1}}{\ddot{\vec{#2}}}}

\newcommand{\vech}[2][ZZZZ]{\ifthenelse{\equal{#1}{ZZZZ}}{}{\presup{#1}}{\mathitbf{\tilde{#2}}}}
\newcommand{\vecb}[2][ZZZZ]{\ifthenelse{\equal{#1}{ZZZZ}}{}{\presup{#1}}{\bar{\underline #2}}}
\newcommand{\mat}[2][ZZZZ]{\ifthenelse{\equal{#1}{ZZZZ}}{}{\presup{#1}\,}{{\boldsymbol #2}}}
\newcommand{\hmat}[2][ZZZZ]{\ifthenelse{\equal{#1}{ZZZZ}}{}{\presup{#1}\,}{{\hat{\boldsymbol #2}}}}
\newcommand{\dmat}[2][ZZZZ]{\ifthenelse{\equal{#1}{ZZZZ}}{}{\presup{#1}\,}{\dot{\boldsymbol #2}}}
\newcommand{\emat}[2][ZZZZ]{\ifthenelse{\equal{#1}{ZZZZ}}{}{\presup{#1}\,}{\hat{\boldsymbol #2}}}
\newcommand{\matfn}[3][ZZZZ]{\ifthenelse{\equal{#1}{ZZZZ}}{}{\presup{#1}}{{\mat{#2}}\left(#3\right)}}
\newcommand{\Rt}[2][ZZZZ]{\ifthenelse{\equal{#1}{ZZZZ}}{}{\presup{#1}}{{\bf R}\left(#2\right)}}

\newcommand{\point}[2][ZZZZ]{\ifthenelse{\equal{#1}{ZZZZ}}{}{\presup{#1}}{\mathbf{\mathrm{#2}}}}
\renewcommand{\frame}[3][ZZZZ]{\ifthenelse{\equal{#1}{ZZZZ}}{}{\presup{#1}}{\mat{#2}}_{#3}}
\newcommand{\frameh}[3][ZZZZ]{\ifthenelse{\equal{#1}{ZZZZ}}{}{\presup{#1}}{\hat{#2}}_{#3}}
\newcommand{\frameb}[3][ZZZZ]{\ifthenelse{\equal{#1}{ZZZZ}}{}{\presup{#1}}{\bar{#2}}_{#3}}

\newcommand{\pnt}[2][ZZZZ]{\ifthenelse{\equal{#1}{ZZZZ}}{}{\presup{#1}}{\mathbf{#2}}}

\newfont{\School}{pncr}
\newfont{\eightTR}{pncr at 8pt}

\usepackage{color}

\author{Dorian Tsai$^{1}$, Donald G. Dansereau$^{2}$, Steve Martin$^{1}$ and Peter Corke$^{1}$}

\newcommand{\QUTtitle}{Mirrored Light Field Video Camera Adapter}
\newcommand{\QUTdocument}{Tech Report}
\newcommand{\QUTrevision}{Rev: 0.5}
\newcommand{\QUTauthor}{Dorian Tsai$^{1}$}
\newcommand{\QUTreviewer}{Donald G. Dansereau$^{2}$, Steve Martin$^{1}$ and Peter Corke$^{1}$}

\begin{document}

\begin{titlepage}
\vspace{-1.5cm}
\begin{center}

\begin{figure*}[!h]
  \centering
  \includegraphics[width=0.2\textwidth]{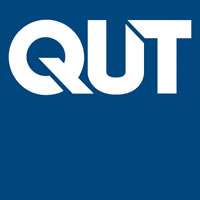}
    \label{fig:QUTlogo}
\end{figure*}

\vspace{1cm}

\textsc{\LARGE Queensland University of Technology}\\

\textsc{\large Faculty of Science and Engineering}

\vspace{2cm}
\textbf{\textsc{\huge \QUTtitle}}
\vspace{1cm}\\
\textbf{\large \QUTdocument}

\vspace{1cm}
\textbf{\large \QUTauthor}\\
\textbf{\large \QUTreviewer}\\
{\small \QUTrevision}

\end{center}
\vfil
\begin{center}
\today\\
{\small Brisbane, Queensland}\\
\vspace{1cm}
{$^{1}$D. Tsai, S. Martin and  P. Corke are with the Australian Centre for Robotic Vision (ACRV), Queensland University of Technology (QUT), Brisbane, Australia {\tt\small \{dy.tsai, steve.martin, peter.corke\}@qut.edu.au}}%

{$^{2}$D. Dansereau is with the Stanford Computational Imaging Lab, Stanford University, CA, USA.  {\tt\small donald.dansereau@gmail.com}}%
\end{center}

\end{titlepage}

\pagestyle{fancy}

\pagenumbering{gobble} 

\hfil
\vspace{3cm}
\begin{abstract}
This paper proposes the design of a custom mirror-based light field camera adapter that is cheap, simple in construction, and accessible. 
Mirrors of different shape and orientation reflect the scene into an upwards-facing camera to create an array of virtual cameras with overlapping field of view at specified depths, and deliver video frame rate light fields. We describe the design, construction, decoding and calibration processes of our mirror-based light field camera adapter in preparation for an open-source release to benefit the robotic vision community.

The latest report, computer-aided design models, diagrams and code can be obtained from the following repository:\\
\url{https://bitbucket.org/acrv/mirrorcam}.
\end{abstract}

\pagebreak


\newpage
\pagenumbering{arabic}
\setcounter{page}{1}

\section{Introduction}

Light field cameras are a new paradigm in imaging technology that may greatly augment the computer vision and robotics fields. Unlike conventional cameras that only capture spatial information in 2D, light field cameras capture both spatial and angular information in 4D using multiple views of the same scene within a single shot~\cite{ng2005light}. Doing so implicitly encodes geometry and texture, and allows for depth extraction. Capturing multiple views of the same scene also allows light field cameras to handle occlusions~\cite{walter2015glossy}, and non-Lambertian (glossy, shiny, reflective, transparent) surfaces, that often break most modern computer vision and robotic techniques~\cite{vaish2006reconstructing}.

\begin{figure}
\centering
\setlength\tabcolsep{2pt}
\begin{tabular}[t]{cc}
	\multirow{2}{*}[28mm]{	
	\subfloat[]{\includegraphics[width=0.5\textwidth,height=0.5\textwidth]{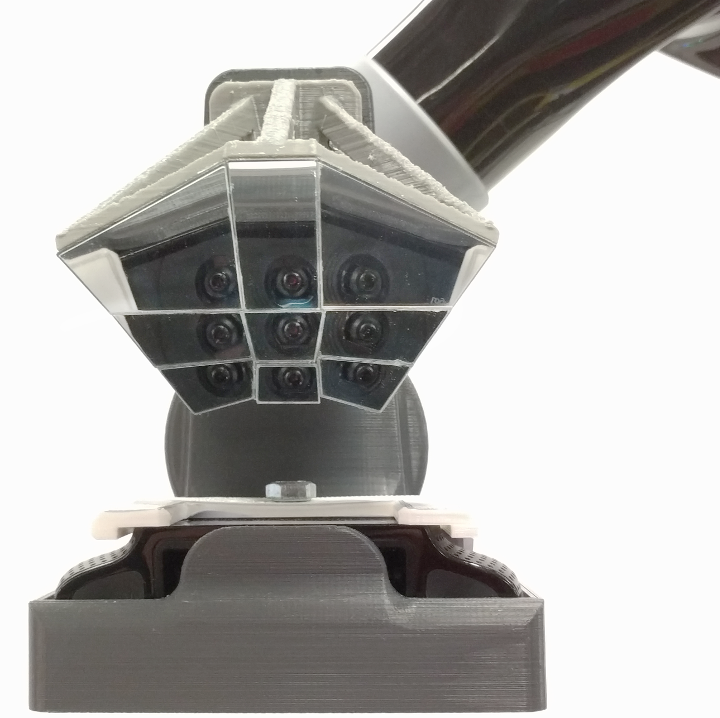} \label{fig:MirrorCam}}
	} & \subfloat[]{\includegraphics[width=0.4\textwidth]{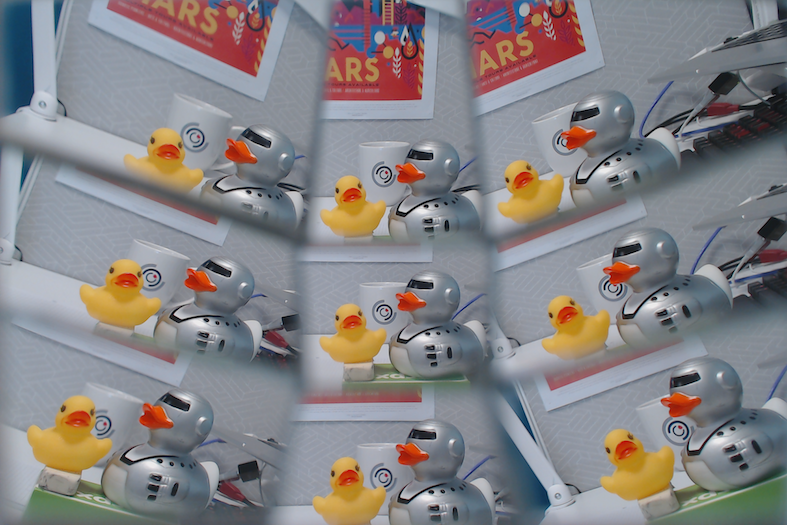}\label{fig:MirrorCamExampleImageA}}\\
	& \subfloat[]{\includegraphics[width=0.4\textwidth]{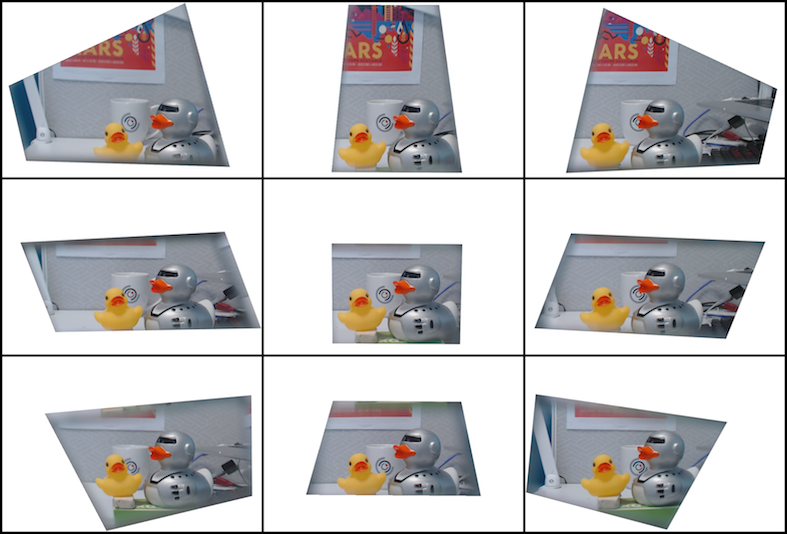}\label{fig:MirrorCamExampleImage}}\\
\end{tabular}
\caption{(a) MirrorCam mounted on the Kinova MICO robot manipulator. Nine mirrors of different shape and orientation reflect the scene into the upwards-facing camera to create 9 virtual cameras, which provides video frame-rate light fields. (b) A whole image captured by the MirrorCam and (c) the same decoded into a light field parameterization of 9 sub-images, visualized as a 2D tiling of 2D images. The non-rectangular sub-images allow for greater \gls{FOV} overlap~\cite{tsai2016lfvs}.}
\label{fig:MirrorCamOverview}
\end{figure} 

Robots must operate in continually changing environments on relatively constrained platforms. As such, the robotics community is interested in low cost, computationally inexpensive, and real-time camera performance. 
Unfortunately, there is a scarcity of commercially available light field cameras appropriate for robotics applications. Specifically, no commercial camera delivers 4D light fields at video frame rates\footnote{Though one manufacturer provides video, it does not provide a 4D LF, only 2D, RGBD or raw lenslet images with no method for decoding to 4D.}. 
Creating a full camera array comes with more synchronization, bulk, input-output and bandwidth issues. However, the advantages of our approach are video-framerate \gls{LF} video allowing real-time performance, the ability to customize the design to optimize key performance metrics required for the application, and the ease of fabrication. The main disadvantages of our approach are a lower resolution, a lower \gls{FOV}\footnote{A $3 \times 3$ array will have $1/3$ the \gls{FOV} of the base camera.}, and a more complex decoding process. 

Therefore, we constructed our own \gls{LF} video camera by employing a mirror-based adapter. This approach splits the camera's field of view into sub-images using an array of planar mirrors. By appropriately positioning the mirrors, a grid of virtual views with overlapping fields of view can be constructed, effectively capturing a light field. We 3D-printed the mount based on our design, and populated the mount with laser-cut acrylic mirrors. 


The main contribution of this paper is
the design and construction of a mirror-based adapter like the one shown in Fig.~\ref{fig:MirrorCam}, which we refer to as MirrorCam.
We provide a novel optimization routine for the design of the custom mirror-based camera that models each mirror using a 3-Degree-of-Freedom (DOF) reflection matrix. 
The calibration step uses 3-DOF mirrors as well; the design step allows non-rectangular projected images. 
We aim to make the design, methodology and code open-source to benefit the robotic vision research community.

The remainder of this paper is organized as follows. Section~\ref{sec:background} provides some background on light field cameras in relation to the MirrorCam. Section~\ref{sec:methods} explains our methods for designing, optimizing, constructing, decoding and calibrating the MirrorCam. 
And finally in Section~\ref{sec:conclusions}, we conclude the paper and explore future work.

\section{Background}
\label{sec:background}

Light field cameras measure the amount of light travelling along each ray that intersects the sensor by acquiring multiple views of a single scene. Doing so allows these cameras to obtain both geometry, texture, and depth information within a single light field image/photograph. Some excellent references for light fields are~\cite{adelson2002single,chan2014lightfield,dansereau2014Thesis}. 

Table~\ref{tbl:cameras} compares some of the most common LF camera architectures. The most prevalent are the camera array~\cite{wilburn2005high}, and the micro-lens array (MLA)~\cite{ng2005light}. However, the commercially-available light field cameras are insufficient for providing light fields for real-time robotics. Notably, the Lytro Illum does not provide light fields at a video frame rate~\cite{lytro2015illumManual}. The Raytrix R10 is a light field camera that captures the light field at more than 7-30 frames-per-second (FPS); however, the camera uses lenslets with different focal lengths, which makes decoding the raw image extremely difficult, and only provides 3D depth maps~\cite{raytrix}.
Furthermore, as commercial products, the light field camera companies have not disclosed details on how to access and decode the light field camera images, forcing researchers to hack solutions with limited success. All of these reasons motivate a customizable, easy-to-access, easy to construct, and open-source video frame-rate light field camera.

\ctable[%
 caption = {Comparison of Accessibility for Different LF Camera Systems},
 label = tbl:cameras,
 width = \textwidth,
 pos = t!,
 star
 ]
 {lcccc}{
 \tnote[1]{Frames per second}
 }
 { \FL \textbf{LF Systems} & \textbf{Sync}  & \textbf{FPS}\tmark[1] & \textbf{Customizability} & \textbf{Open-Source} \ML
  Camera Array 		& poor 	& 7-30 	& significant	& yes \\
  MLA (Lytro Illum) & good 	& 0.5 	& none 			& limited  	\\
  MLA (Raytrix R8/R10)			& good	& 7-30	& minor			& limited 	\\
  MirrorCam	& good	& 2-30 	& significant 	& yes \LL
 }


\section{Methods}
\label{sec:methods}

We constructed our own \gls{LF} video camera by employing a mirror-based adapter based on previous works~\cite{fuchs2013design,song2015light,mukaigawa2010hemispherical}. 
This approach slices the original camera image into sub-images using an array of planar mirrors. Curved mirrors may produce better optics; however, these mirrors are difficult to produce. Planar mirrors are much more accessible and customizable. A grid of virtual views with overlapping field of view can be constructed by carefully aligning the mirrors. These multiple views effectively capture a light field.

Our approach differs from previous work by reducing the optimization routine to a single tunable parameter, and identifying the fundamental trade-off between depth of field and field of view in the design of mirrored LF cameras. Additionally, we utilize non-square mirror shapes.

\subsection{Design \& Optimization}
\label{subsec:design}


Because an array of mirrors has insufficient degrees of freedom to provide both perfectly overlapping \glspl{FOV} and perfectly positioned projective centres, we employ an optimization algorithm to strike a balance between these factors, as in~\cite{fuchs2013design}. A tunable parameter determines the relative importance of closeness to a perfect grid of virtual poses, and field of view overlap, which is evaluated at a set of user-defined depths. The grid of virtual poses is allowed to be rectangular, to better exploit rectangular camera \glspl{FOV}. 

The optimization routine begins with a faceted parabola at a user-defined scale and mirror count. Optimization is allowed to manipulate the positions and normals of the mirror planes, as well as their extents. Optimization constraints prevent mirrors occluding their neighbours, and allow a minimum spacing between mirrors to be imposed for manufacturability.

\newsavebox{\MirrorCamSbox}
\begin{figure}
	\centering
	\sbox{\MirrorCamSbox}{\includegraphics[width=0.7\columnwidth]{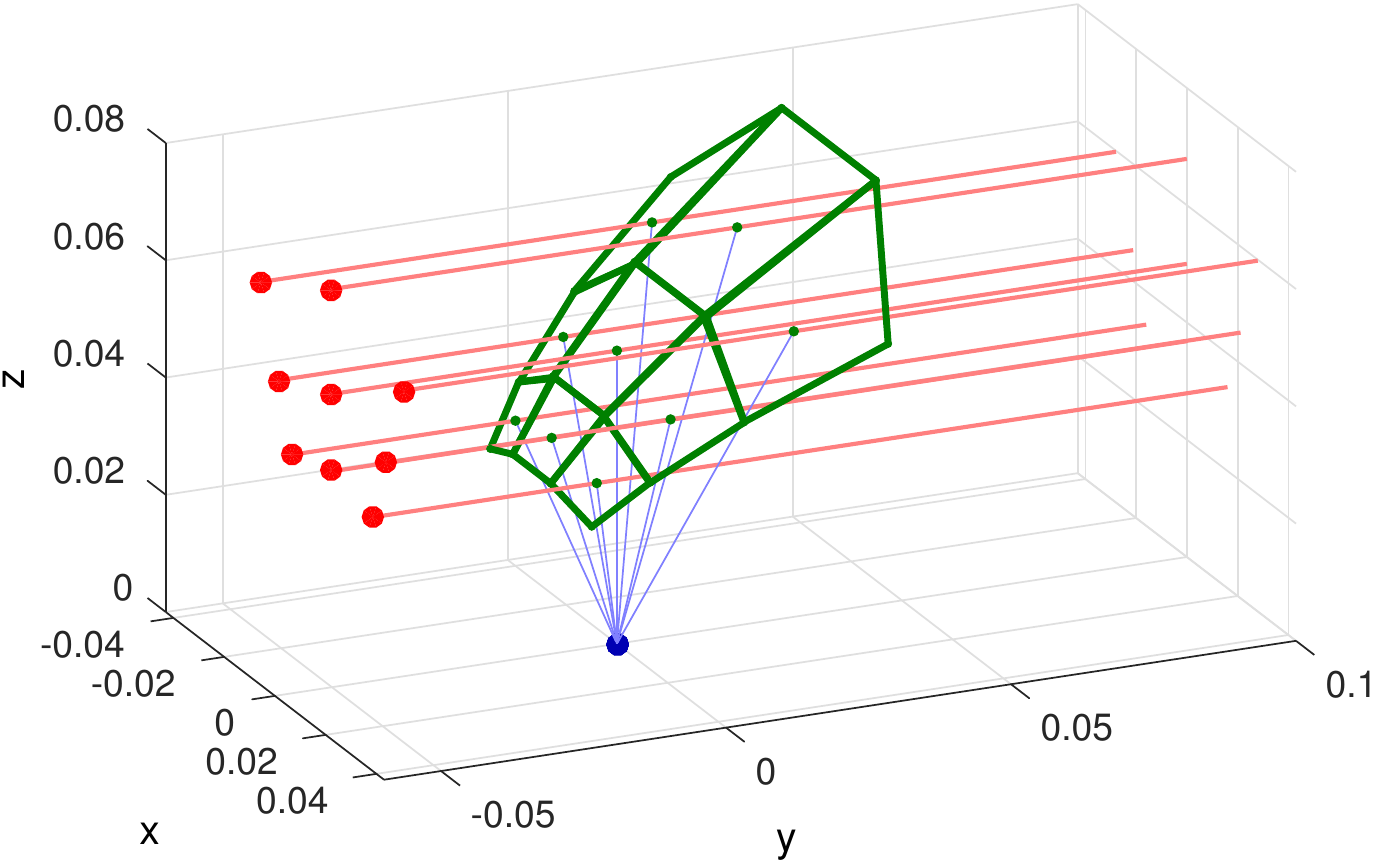}}
	\subfloat[]{\usebox{\MirrorCamSbox}}
	\subfloat[]{\vbox to \ht\MirrorCamSbox{\vfil\hbox{\includegraphics[width=0.3\columnwidth]{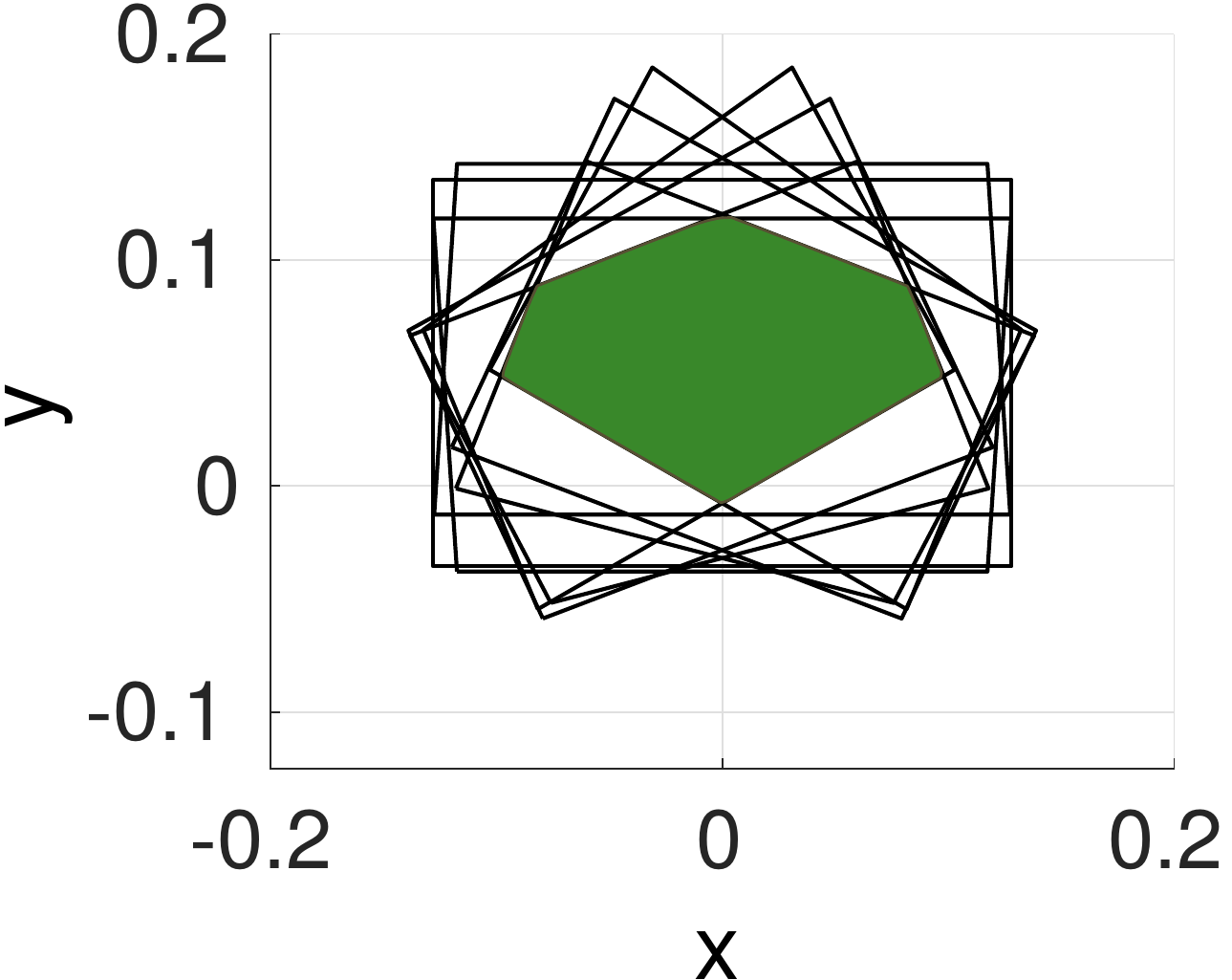}}\vfil}}\\
	\subfloat[]{\includegraphics[width=0.7\columnwidth]{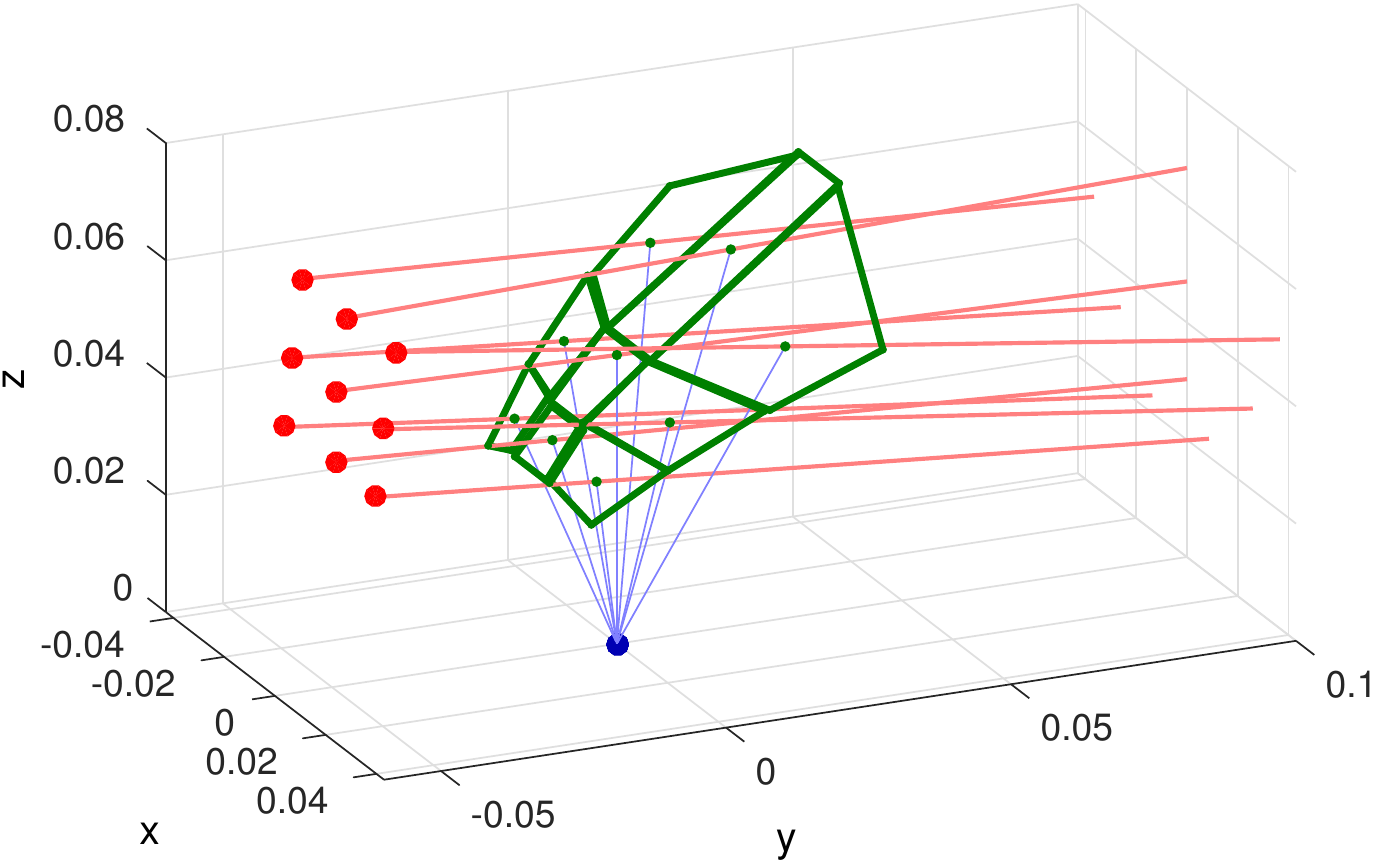}}
	\subfloat[]{\vbox to \ht\MirrorCamSbox{\vfil\hbox{\includegraphics[width=0.3\columnwidth]{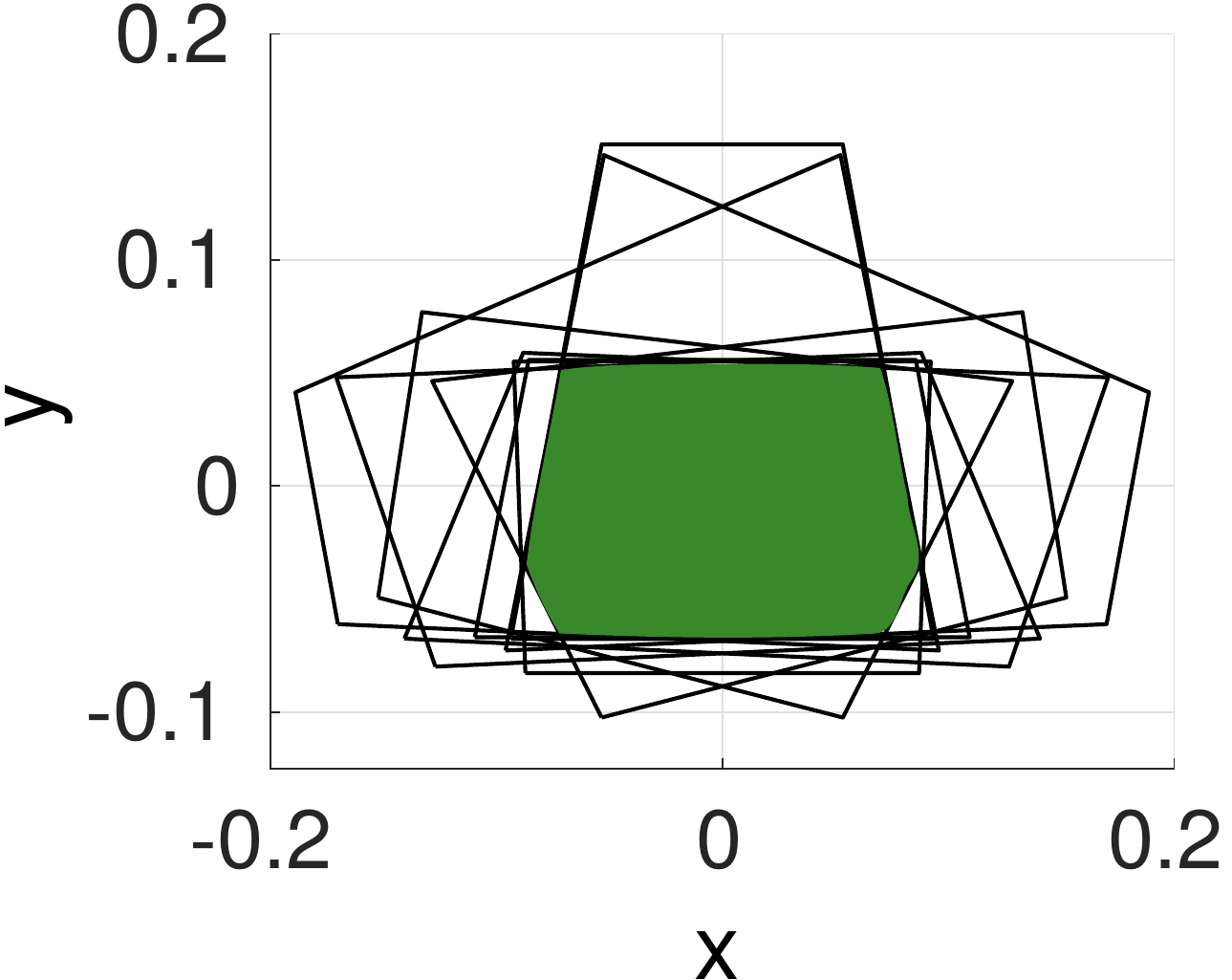}}\vfil}}	
\caption{(a) A parabolic mirror array reflects images from the scene at right into a camera, shown in blue at bottom; Each mirror yields a virtual view, shown in red --~note that these are far from an ideal grid; (b) The \gls{FOV} overlap evaluated at 0.5~m, with the region of full overlap highlighted in green; (c) and (d) the same after optimization, showing better virtual camera placement and \gls{FOV} overlap.}
\label{fig:OptimizingMirrorCam}
\end{figure} 

Fig.~\ref{fig:OptimizingMirrorCam} shows an example $3 \times 3$ mirror array before and after optimization. The \gls{FOV} overlap was evaluated at 0.3 and 0.5~m. Fig.~\ref{fig:MirrorCam} shows an assembled model mounted on a robot arm, and Fig.~\ref{fig:MirrorCamExampleImageA} shows an example image taken from the camera. Note that the optimized design does not yield rectangular sub-images, as allowing a general quadrilateral shape allows for greater \gls{FOV} overlap. In future work, we will explore the use of non-quadrilateral sub-images.

\subsection{Construction}
\label{subsec:construction}

\begin{figure}
\centering
\subfloat[]{\includegraphics[height=0.5\textwidth]{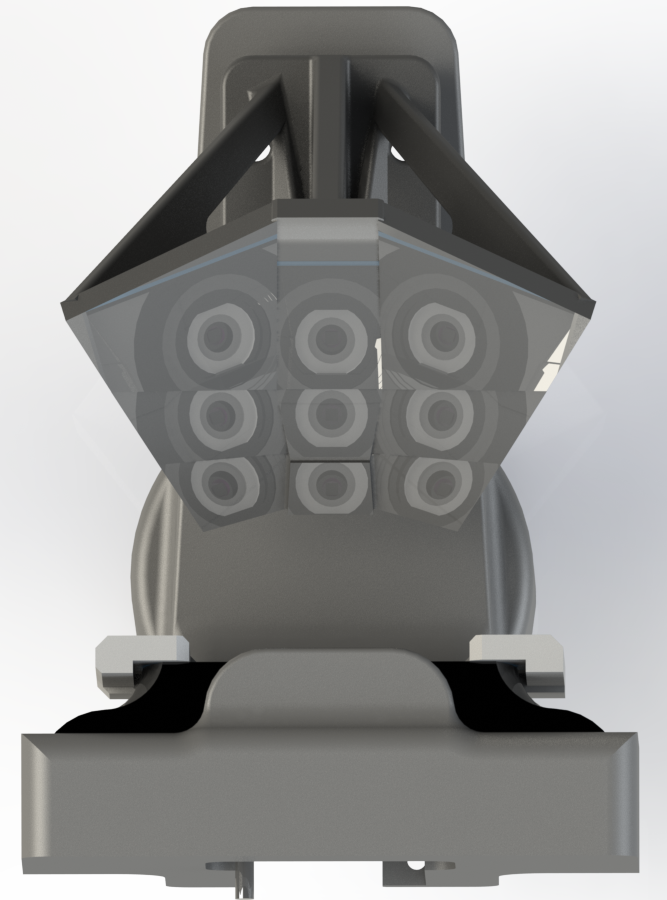}\label{fig:mirrorCamRender04_1}}\hfil
\subfloat[]{\includegraphics[height=0.5\textwidth]{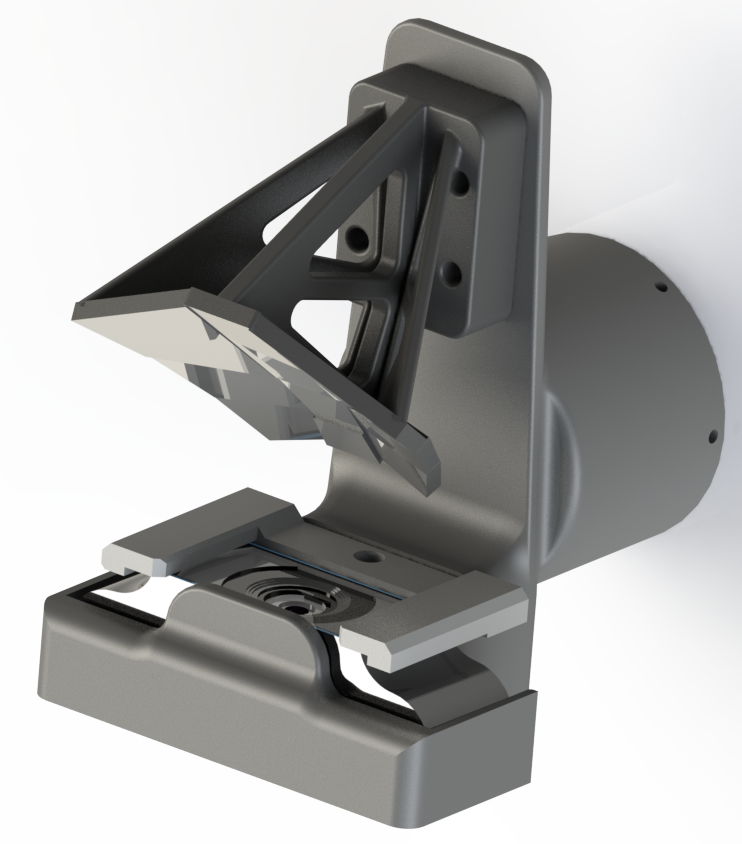} \label{fig:mirrorCamRender04_2.}}
\caption{Rendered image of the MirrorCam version 0.4C, (a) from the front showing the single camera lens that is visible from all nine mirrored surfaces, and (b) an isometric view showing how the camera is attached to the mirrors.}
\label{fig:mirrorCamRender}
\end{figure}

For the construction of the MirrorCam, we aimed to use easily accessible materials and methods. We 3D-printed the mount based on our design, and populated the mounts with laser-cut flat acrylic mirrors. Figure~\ref{fig:mirrorCamRender} shows a computer rendering of the MirrorCam before 3D printing. The reflection of the 9 mirrors show the upwards-facing camera, which is secured at the base of the MirrorCam. This design was built for the commonly available Logitech C920 webcam. More detailed diagrams of the design are supplied in the Appendix.


Mirror thickness and quality proved to be an issue for the construction of the MirrorCam. Since the mirrors are quite close to the camera, the thickness of the mirrors occlude a significant portion of the image, which greatly reduces the resolution of each sub-image. Thus, we opted for thin mirrors, but encountered problems with mirror warping and flatness from the cheap acrylic mirrors. By inspecting the mirrors before purchase, and handling them very carefully (without flexing them) during construction, cutting and adhesion, we were able to minimise image warping and flatness.

\subsection{Decoding \& Calibration}
\label{subsec:calibration}


Our MirrorCam calibration has two steps: first the base camera is calibrated following a conventional intrinsic calibration, e.g. using MATLAB's built-in camera calibration tool. Next the camera is assembled with mirrors and the mirror geometry is estimated using a Levenberg-Marquardt optimization of the error between expected and observed checker board corner locations. Initialization of the mirror geometry is based on the array design, and sub-image segmentation is manually specified.

One point of difference with prior work is that rather than employing a 6-\gls{DOF} transformation for each virtual camera view, our calibration models each mirror using a 3-\gls{DOF} reflection matrix. 
This reduces the DOF in the camera model and more closely matches the physical camera, speeding convergence and improving robustness.

A limitation of our calibration technique is that the images taken without mirrors are only considered when initializing the camera intrinsics. A better solution, left as future work, would jointly consider all images, with and without mirrors.

Based on the calibrated mirror geometry, the nearest grid of parallel cameras is estimated, and decoding proceeds as:
\begin{enumerate}
\item Remove 2D radial distortion,
\item Slice 2D image into a 4D array, and
\item Reproject each 2D sub-image into central camera view orientation.
\end{enumerate}
Here, we assume the central camera view is aligned with the center mirror.

The final step corrects for rotational differences between the calibrated and desired virtual camera arrays using 2D projective transformations. There is no compensation for translational error, though in practice the cameras are very close to an ideal grid. An example input image and decoded light field are shown in Fig.~\ref{fig:MirrorCamExampleImage}. Our calibration routine reported a 3D spatial reprojection \gls{RMS} error of 1.80~mm. The spatial reprojection error is the 3D distance from the projected ray to the expected feature location during camera calibration, where pixel projections are traced through the camera model into space. This small error confirms that the camera design, manufacture and calibration has yielded observations close to an ideal light field. 

It is important to note that our current calibration did not account for the manufacturing aspects of the camera, such as the thickness of the acrylic mirrors, or the additional thickness of the epoxy used to secure the mirrors to the mount. The acrylic mirrors we used also exhibited some bending and rippling, causing image distortion unaccounted for in the calibration process.


\subsubsection{Light Field Features}
\label{subsec:features}

As detailed in our report on light field image-based visual servoing~\cite{tsai2016lfvs}, to our knowledge all prior work on light field features operates by applying 2D methods to epipolar slices of the 4D light field.  As a first step towards truly 4D features, we augment both the detection and representation of features to exploit the light field structure. Our implementation employs Speeded-Up Robust Features (SURF)~\cite{bay2008surf}, though the proposed method is agnostic to feature type. We augment the detection process by enforcing light field geometry constraints, making the process more selective and less sensitive to spurious detections.  Finally, for representation, we augment the 2D feature location with the local light field slope, implicitly encoding depth. 

\subsubsection{The Point-Plane Correspondence}

Prior work has extracted light field features by applying 2D feature detectors to 2D slices in the $u,v$ dimensions~\cite{johannsen2015linear}. In this paper, we do the same, but with an additional step to reject features that break the point-plane correspondence. By selecting only features that adhere to the planar relationship, we can remove spurious detections and limit our attention to features that correspond consistently to 3D spatial locations. 

Operating on 2D slices of the light field, feature matches are found between the central view and all other sub-images. Each pair of matched 2D features is treated as a potential 4D feature. A single feature pair yields a slope estimate, and this defines an expected feature location in all other sub-images. We introduce a tunable constant that determines the maximum distance between observed and expected feature locations, in pixels, and reject all matches exceeding this limit.  

A second constant $N\sub{MIN}$ imposes the minimum number of sub-images in which feature matches must be found.  In the absence of occlusions, this can be set to require feature matches in all sub-images. Any feature passing the maximum distance criterion in at least $N\sub{MIN}$ images is accepted as a 4D feature, and a consolidated slope estimate is formed based on all passing sub-images. The result of this decoding process is a light field ready for experimental validation.

\section{Conclusions and Future Work}
\label{sec:conclusions}

In this paper, we have proposed the design optimisation, construction, decoding and calibration process of a mirror-based light field camera. We have shown that our 3D-printed MirrorCam, optimized for overlapping FOV, reproduced a light field. 
This implies that the mirror-based \gls{LF} camera was a viable, low-cost, and accessible alternative to commercially available \gls{LF} cameras. 

Our implementation takes 5 seconds per frame to operate as unoptimized MATLAB code. The decoding and correspondence processes are the current bottlenecks. Through optimization, real-time light fields should be possible. We push the envelope of technology towards real-time light field cameras for robotics.

In future work, we will validate the MirrorCam in terms of image refocusing, depth estimation and perspective shift in comparison to other commercially-available light field cameras.




\section*{Acknowledgement}

This research was partly supported by the Australian Research Council (ARC) Centre of Excellence for Robotic Vision (CE140100016). We also thank the other members of the ACRV for their insight and guidance.



\bibliographystyle{IEEEtran}
\bibliography{IEEEabrv,LightFieldBib}

\begin{thebibliography}{10}
\providecommand{\url}[1]{#1}
\csname url@rmstyle\endcsname
\providecommand{\newblock}{\relax}
\providecommand{\bibinfo}[2]{#2}
\providecommand\BIBentrySTDinterwordspacing{\spaceskip=0pt\relax}
\providecommand\BIBentryALTinterwordstretchfactor{4}
\providecommand\BIBentryALTinterwordspacing{\spaceskip=\fontdimen2\font plus
\BIBentryALTinterwordstretchfactor\fontdimen3\font minus
  \fontdimen4\font\relax}
\providecommand\BIBforeignlanguage[2]{{%
\expandafter\ifx\csname l@#1\endcsname\relax
\typeout{** WARNING: IEEEtran.bst: No hyphenation pattern has been}%
\typeout{** loaded for the language `#1'. Using the pattern for}%
\typeout{** the default language instead.}%
\else
\language=\csname l@#1\endcsname
\fi
#2}}

\bibitem{ng2005light}
R.~Ng, M.~Levoy, M.~Bredif, G.~Duval, M.~Horowitz, and P.~Hanrahan, ``Light
  field photography with a hand-held plenoptic camera,'' Stanford University
  Computer Science, Tech. Rep., 2005.

\bibitem{walter2015glossy}
C.~Walter, F.~Penzlin, E.~Schulenburg, and N.~Elkmann, ``Enabling multi-purpose
  mobile manipulators: Localization of glossy objects using a light-field
  camera,'' in \emph{Conference on Emerging Technologies \& Factory Automation
  (ETFA)}.\hskip 1em plus 0.5em minus 0.4em\relax IEEE, 2015, pp. 1--8.

\bibitem{vaish2006reconstructing}
V.~Vaish, M.~Levoy, R.~Szeliski, C.~Zitnick, and S.~Kang, ``Reconstructing
  occluded surfaces using synthetic apertures: Stereo, focus and robust
  measures,'' in \emph{Intl. Conference on Computer Vision and Pattern
  Recognition (CVPR)}, vol.~2.\hskip 1em plus 0.5em minus 0.4em\relax IEEE,
  2006, pp. 2331--2338.

\bibitem{tsai2016lfvs}
D.~Tsai, D.~G. Dansereau, T.~Peynot, and P.~Corke, ``image-based visual
  servoing with light field cameras,'' \emph{Robotics and Automation Letters},
  2016 [submitted].

\bibitem{adelson2002single}
E.~H. Adelson and J.~Y.~A. Wang, ``Single lens stereo with a plenoptic
  camera,'' \emph{IEEE Transactions on Pattern Analysis and Machine
  Intelligence (TPAMI)}, vol.~14, no.~2, pp. 99--106, 2002.

\bibitem{chan2014lightfield}
S.~C. Chan, ``Light field,'' in \emph{Computer Vision A Reference Guide},
  K.~Ikeuchi, Ed.\hskip 1em plus 0.5em minus 0.4em\relax Springer Link, 2014,
  pp. 447--453.

\bibitem{dansereau2014Thesis}
D.~G. Dansereau, ``Plenoptic signal processing for robust vision in field
  robotics,'' Ph.D. dissertation, University of Sydney, Jan. 2014.

\bibitem{wilburn2005high}
B.~Wilburn, N.~Joshi, V.~Vaish, E.~Talvala, E.~Antunez, A.~Barth, A.~Adams,
  M.~Horowitz, and M.~Levoy, ``High performance imaging using large camera
  arrays,'' \emph{ACM Transactions on Graphics (TOG)}, vol.~24, no.~3, pp.
  765--776, 2005.

\bibitem{lytro2015illumManual}
Lytro, \emph{Lytro Illum User Manual}, Lytro Inc., Mountain View, CA, 7 2015.

\bibitem{raytrix}
\BIBentryALTinterwordspacing
Raytrix, ``Raytrix light field sdk,'' Ratrix GmbH, 2015. [Online]. Available:
  \url{https://www.raytrix.de/Rx.ApiLF.3.1/}
\BIBentrySTDinterwordspacing

\bibitem{fuchs2013design}
M.~Fuchs, M.~K{\"a}chele, and S.~Rusinkiewicz, ``Design and fabrication of
  faceted mirror arrays for light field capture,'' in \emph{Computer Graphics
  Forum}, vol.~32, no.~8.\hskip 1em plus 0.5em minus 0.4em\relax Wiley Online
  Library, 2013, pp. 246--257.

\bibitem{song2015light}
W.~Song, Y.~Liu, W.~Li, and Y.~Wang, ``Light field acquisition using a planar
  catadioptric system,'' \emph{Optics Express}, vol.~23, no.~24, pp.
  31\,126--31\,135, 2015.

\bibitem{mukaigawa2010hemispherical}
Y.~Mukaigawa, S.~Tagawa, J.~Kim, R.~Raskar, Y.~Matsushita, and Y.~Yagi,
  ``Hemispherical confocal imaging using turtleback reflector,'' in
  \emph{Computer Vision--ACCV 2010}.\hskip 1em plus 0.5em minus 0.4em\relax
  Springer, 2010, pp. 336--349.

\bibitem{bay2008surf}
H.~Bay, A.~Ess, T.~Tuytelaars, and L.~V. Gool, ``Speeded-up robust features
  ({SURF}),'' \emph{Computer Vision and image understanding}, vol. 110, no.~3,
  pp. 346--359, 2008.

\bibitem{johannsen2015linear}
O.~Johannsen, A.~Sulc, and B.~Goldluecke, ``On linear structure from motion for
  light field cameras,'' in \emph{Intl. Conference on Computer Vision (ICCV)},
  2015, pp. 720--728.

\end{thebibliography}

\newpage
\clearpage

\appendix

\section{Diagrams}
\begin{figure}[h]
\centering
\includegraphics[width=0.95\textwidth,angle=90]{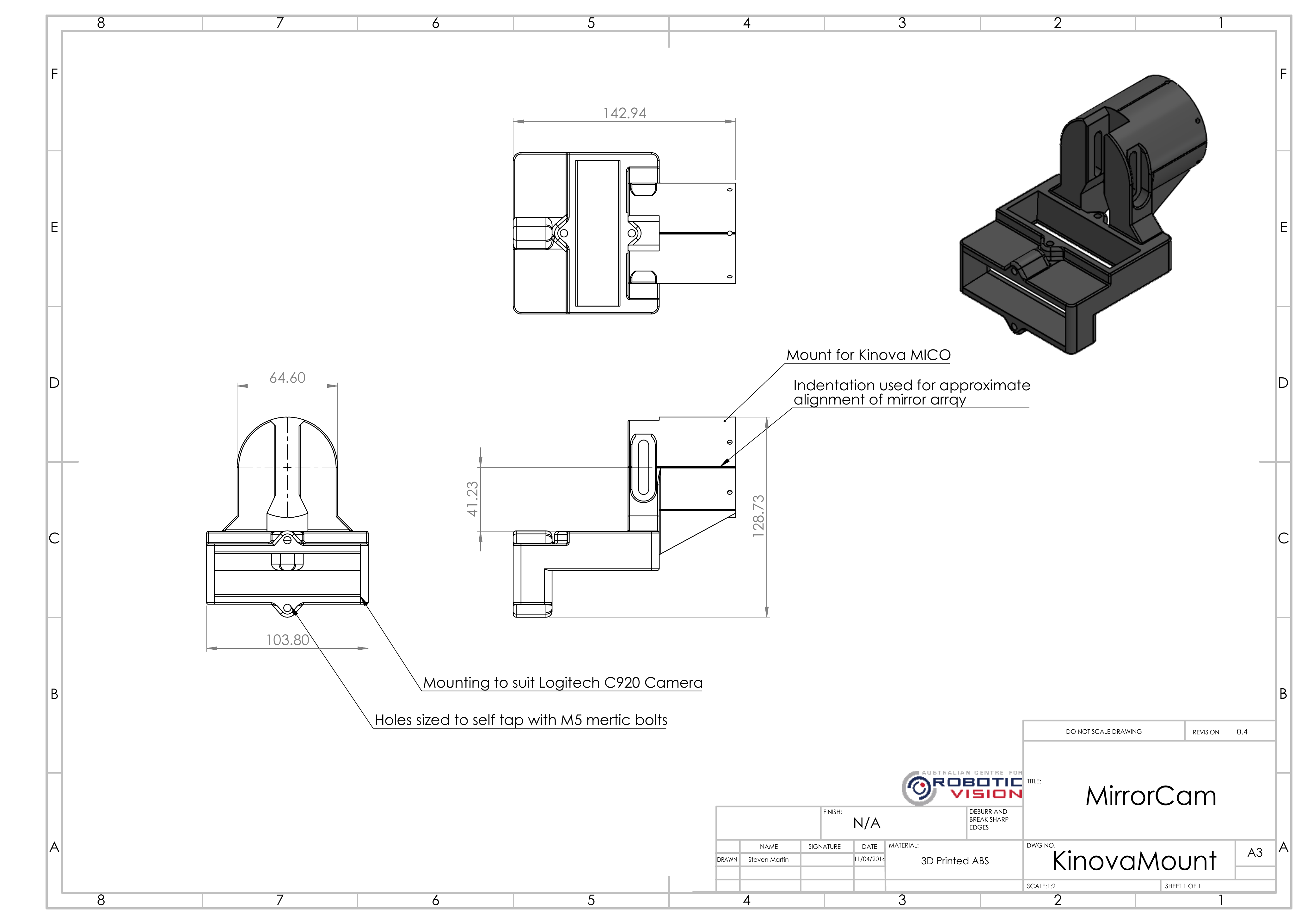}
\caption{MirrorCam v0.4c kinova mount.}
\label{fig:mirrorCamClip1}
\end{figure} 

\begin{figure}
\centering
\includegraphics[width=1.0\textwidth,angle=90]{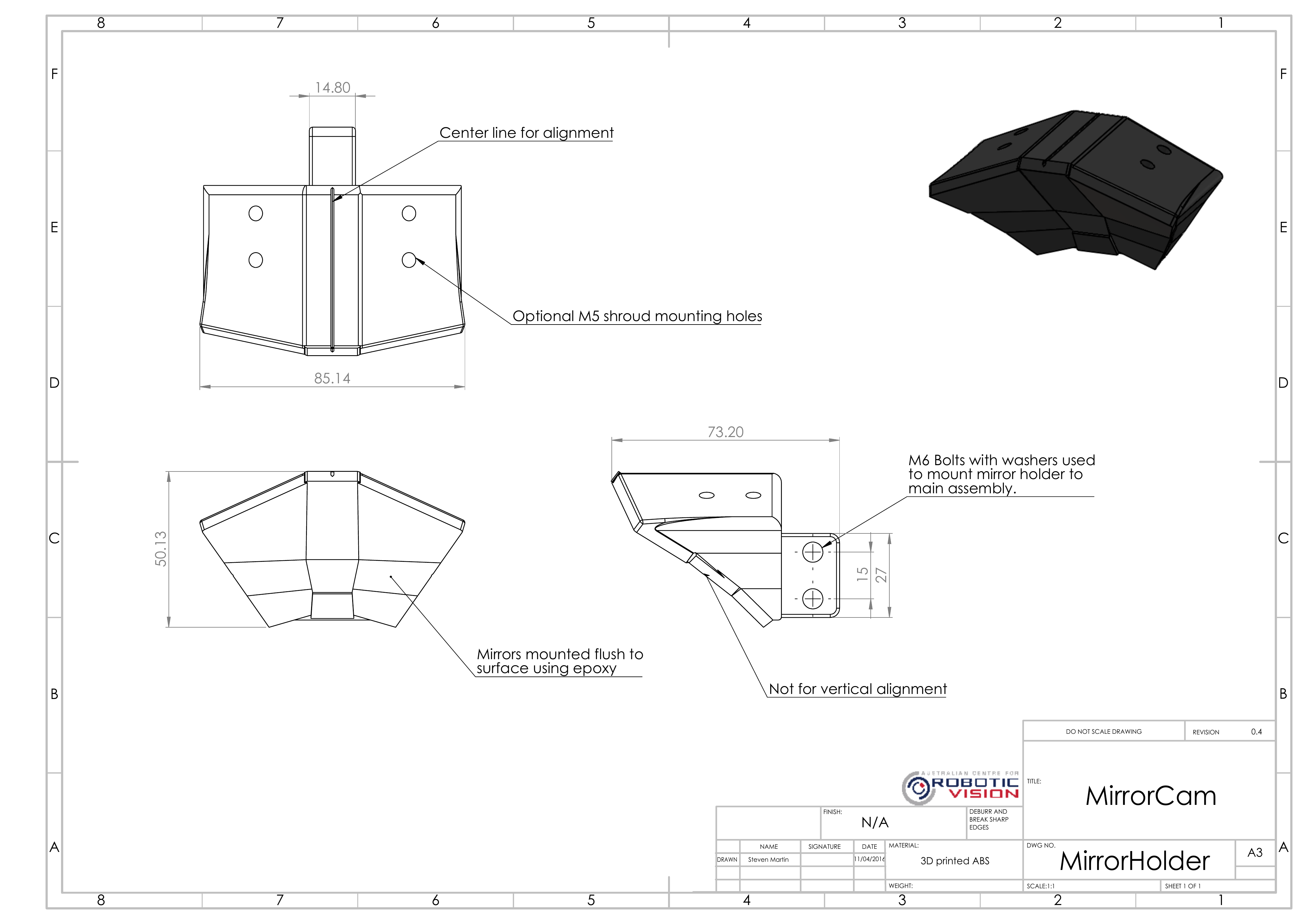}
\caption{MirrorCam v0.4c mirror holder.}
\label{fig:mirrorCamClip2}
\end{figure} 

\begin{figure}
\centering
\includegraphics[width=1.0\textwidth,angle=90]{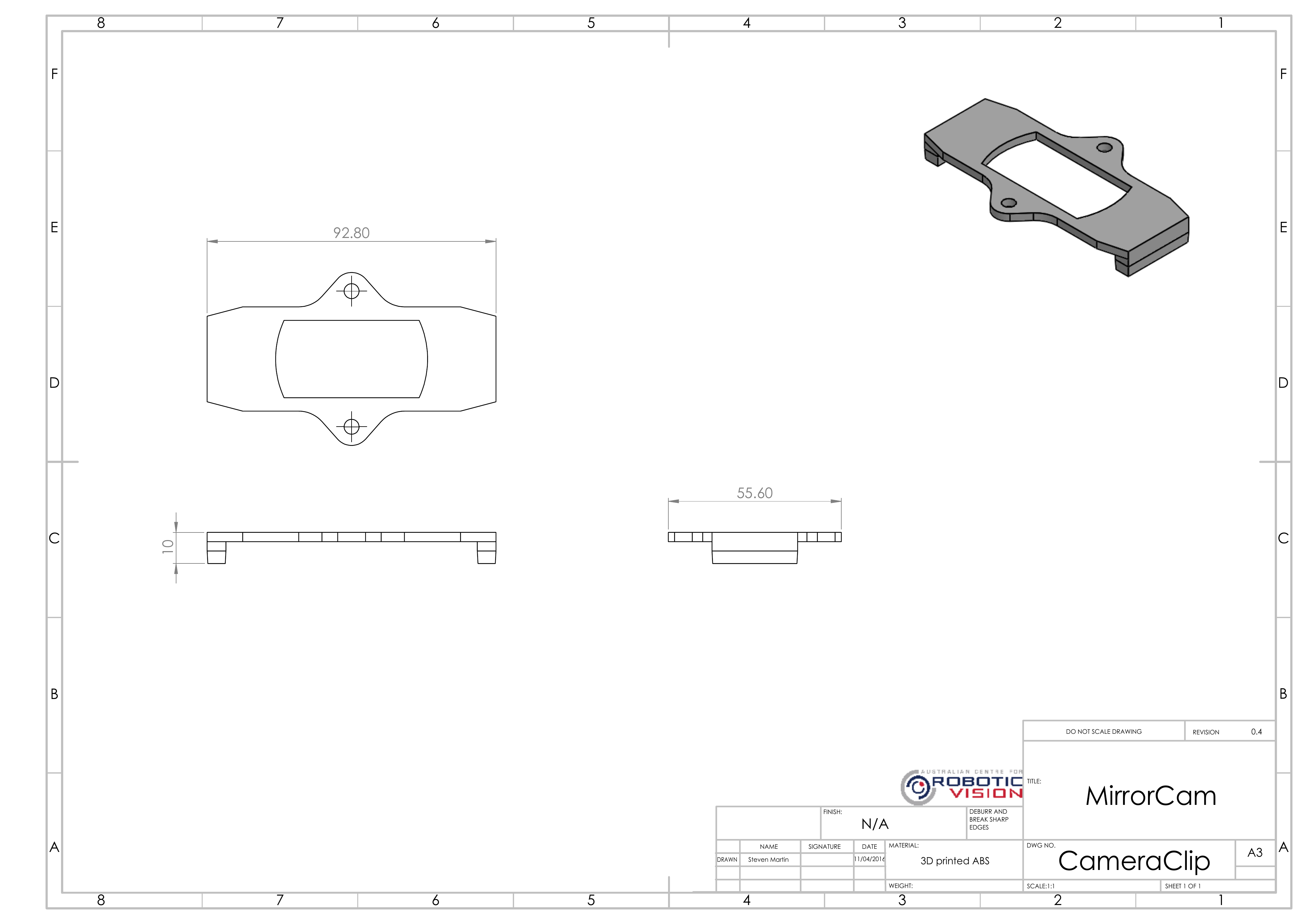}
\caption{MirrorCam v0.4c camera clip.}
\label{fig:mirrorCamClip3}
\end{figure}

\end{document}